# Script Normalization for Unconventional Writing of Under-Resourced Languages in Bilingual Communities

**Sina Ahmadi**          **Antonios Anastasopoulos**
Department of Computer Science
George Mason University
{sahmad46,antonis}@gmu.edu

## Abstract

The wide accessibility of social media has provided linguistically under-represented communities with an extraordinary opportunity to create content in their native languages. This, however, comes with certain challenges in script normalization, particularly where the speakers of a language in a bilingual community rely on another script or orthography to write their native language. This paper addresses the problem of script normalization for several such languages that are mainly written in a Perso-Arabic script. Using synthetic data with various levels of noise and a transformer-based model, we demonstrate that the problem can be effectively remediated. We conduct a small-scale evaluation of real data as well. Our experiments indicate that script normalization is also beneficial to improve the performance of downstream tasks such as machine translation and language identification.[1]

Figure 1: A comparison of the Perso-Arabic scripts used in the selected languages.

## 1 Introduction

With the increasing accessibility of the internet around the globe, from nomad settlements to megacities, we are living in a critical era that gives voice to different language communities to be represented online. Text as the essential material of many applications in natural language processing (NLP) may not be consistently created by under-represented communities due to the lack of a writing system and/or tradition, a widely-used orthography, and reasons related to regional and geopolitical conditions. Moreover, speakers of a language may opt for another script or prefer to use the script of an administratively dominant language other than the conventional one used in their native language (Gathercole and Thomas, 2009). This can be due to illiteracy in the native or home language, diglossia, or simply, the lack of adequate support for specific languages and scripts re-sulting in remarkable negative effects on writing (Oriyama, 2011; Eisenchlas et al., 2013).

Despite the fascinating progress in language technology in recent years, there are many fundamental tasks in NLP that are usually deemed trivial yet are ubiquitous, needed, and *still unsolved* in low-resource settings. One of these tasks is to deal with noisy text which affects the quality and performance of various applications such as text mining (Dey and Haque, 2009) and machine translation (Sperber et al., 2017). In practice, some steps are generally taken to deal with inconsistencies in the text, such as unifying character encoding and rectifying orthographic rules – a task broadly referred to as data cleaning or text preprocessing (Chu et al., 2016). However, unconventional and unsystematic writing such as using the script of a language to write in another language presents challenges beyond traditional data cleaning scenarios.

In this vein, we address the task of script normalization which aims to normalize a text writ-ten in a script or orthography other than the one that is widely used in the language as the conventional one. For instance, writing Kashmiri or Sorani Kurdish in a script like that of Urdu or Persian, rather than in their own conventional scripts. Although this task has much in common with data cleaning, transliteration (Ahmadi, 2019), text normalization as defined by Sproat and Jaitly (2016) and spelling error correction (Hládek et al., 2020), it is particular in a few ways: unconventional writing does not necessarily comply with orthographic rules of the text's language; when writing borrowed or common words from the dominant language, there is an influence of the orthography of the donor language rather than writing in the recipient language's script or orthography; phonemes and graphemes may not be represented according to any specific rule, as in writing /ʃ/ as 'ch', 'sh' or '$' in *Arabizi* (Al-Badrashiny et al., 2014) following an erratic or obscure pattern among speakers leading to a huge amount of noisy material. A few examples of unconventional writing in Kurdish, Kashmiri and Gilaki are shown in Table 1.

---

[1] The data and code are publicly available at https://github.com/sinaahmadi/ScriptNormalization.

| Language | Unconventional script | Unconventional writing | Conventional writing |
|---|---|---|---|
| Gilaki | Persian | یته زون نم هیسه گه گیلکن اون جی گب زن | یته زوؤن ً نؤم هیسه گه گیلکؤن اؤن ً جی گب زن |
| Kashmiri | Urdu | برور چھ اکھ وراے جانور۔ | برؤر چھ ؙ اکھ ُوراے کی جانوؔر۔ |
| Kurmanji | Arabic | قایمقام المدي بةرثوا پارزکار دهوک دا | قایمقام ئامیدیی بەرسڤا پارزیگاری ٘ دهؤکی دا |
| Sorani | Arabic | هقر لة يةکةم شانؤوة ديارة فهديان دةويت | ھەر لە یەکەم شانؤوە دیارە فەھمدیان دەویت |
| Sindhi | Urdu | مدني ڈانهن هجرت وقت فقط حمي ءُ کهروارِي سانڌ گڏ هئي | مديني ڈانهن هجرت وقت فقط حمي، کهرواري سانڌ کڏ هئي |

Table 1: Examples of scripts used unconventionally for writing Gilaki, Kashmiri, Kurdish and Sindhi. The highlighted words contain unconventional characters.

In this paper, we focus on the script normalization of a few under-resourced languages that use variants of the Perso-Arabic script, as schematized in Figure 1, and are spoken as minority languages in bilingual communities with a dominant language using a similar but different script. These languages are Azeri Turkish (AZB), Mazanderani (MZN), Gilaki (GLK), Sorani Kurdish (CKB), Kurmanji Kurdish (KMR), Gorani (HAC), Kashmiri (KAS) and Sindhi (SND). Although these languages have their own customized scripts with more or less defined orthographies in their communities, they are oftentimes written in the script of the dominant language, notably Persian (FAS), Arabic (ARB) and Urdu (URD) scripts. Furthermore, these languages have been lexically and, to a lesser extent, typologically influenced by the administratively dominant languages. Akin to many other multilingual and pluricultural societies, speakers of these languages have faced language conflict and linguistic discrimination in different educational, political, cultural, and communicative domains, and struggle with ethnolinguistic vitality (Mohan, 1989; Shah, 1997; Bird, 2020; Sheyholislami, 2022). Appendix A presents a summary of the languages we study.

**Contributions** In this work, we aim to:

1. shed light on script normalization for under-resourced languages with very limited progress in language technology to facilitate the identification and collection of relevant data in the future,

2. leverage synthetic data for script normalization by mapping scripts based on rules and sequence alignment, and

3. cast script normalization as a translation task where noisy text is "translated" into normalized one using synthetic data generated.

We demonstrate that imposing different levels of noise on the synthetic data is beneficial to train more robust transformer-based models to normalize scripts and also, improve the performance of downstream tasks such as machine translation and language identification of unconventional writing.

## 2 Related Work

Although some aspects of script normalization have been previously addressed in related NLP tasks, its definition is a rather subjective matter, where a set of intentional or unintentional anoma-lies are "normalized". Therefore, script normalization overlaps considerably with more well-defined tasks such as spelling correction, lexical normalization (van der Goot et al., 2021) where an utterance is transformed into its standard form, pattern recognition (Schenk et al., 2009; Maddouri et al., 2000), language identification and standardization (Partanen et al., 2019; Ahmadi, 2020c).

Script and text normalization have been proven beneficial in various downstream applications such as dependency parsing (Zhang et al., 2013), sentiment analysis (Mandal and Nanmaran, 2018) and named-entity recognition (Baldwin and Li, 2015). Although in some contexts normalization has been used to refer to basic tasks such as stemming and lemmatization, as in (Toman et al., 2006), those are not within the scope of this paper.

**Text Normalization** One of the most related tasks to script normalization is text normalization which broadly deals with alternative spellings, typos, abbreviations and non-canonical language and is of importance to text-to-speech systems and for handling micro-blogging data such as Tweets (Sproat and Jaitly, 2016). To this end, a wide range of techniques have been proposed using rules based on annotated corpora (Sigurðardóttir et al., 2021) or linguistic information (Xia et al., 2006), edit operations and recurrent neural networks (Chrupała, 2014), machine translation (Graliński et al., 2006), supervised learning (Yang and Eisenstein, 2013), encoder-decoders (Lusetti et al., 2018) and more recently, transformers (Zhang et al., 2019; Tan et al., 2020; Bucur et al., 2021). MoNoise (van der Goot and van Noord, 2017) is a prominent approach to text normalization where the problem is framed as a domain adaptation one and various steps are taken to generate and rank normalized candidates using spell checkers, word embeddings, dictionaries and $n$-grams features.

**Perso-Arabic Script Normalization** As one of the important scripts adopted by languages spoken by over 600 million speakers (Doctor et al., 2022), the Perso-Arabic scripts are prevalent on the Web nowadays. Although script normalization in general and addressing ambiguities of writing systems, in particular, have been previously addressed in the related tasks for such languages, such as Arabic (Ayedh et al., 2016; Shaalan et al., 2019), Kashmiri (Lone et al., 2022a), Kurdish (Ahmadi, 2019) and Sindhi (Jamro, 2017), normaliz-ing Perso-Arabic scripts has not received much attention, let alone for noisy data originated from unconventional writing in bilingual communities. In a recent study, Doctor et al. (2022) address the normalization of Perso-Arabic script for a number of languages, namely Urdu, Punjabi, Sindhi, Kashmiri, Sorani Kurdish, Uyghur and Azeri Turkish. Inspired by Johny et al. (2021)'s approach to using finite-state transducers (FSTs) to normalize Brahmic scripts, Gutkin et al. (2022) implement FSTs for Perso-Arabic scripts for Unicode normalization, visual normalization and reading normalization by focusing on normalization and unification of varieties based on regional orthographies rather than that of specific dominant scripts.

**Low-resource Setup** Most under-resourced languages that require script normalization face the predicament of data paucity. On the other hand, data annotation is a tedious task that may not be always feasible for all languages. To tackle these, Dekker and van der Goot (2020) create synthetic data in which canonical words are replaced with non-canonical ones. Lusito et al. (2022) use a transformer-based model and employ modern data augmentation techniques for the endangered language of Ligurian; to deal with the scarcity of data, "back-normalization" is proposed where normalized text is transformed to a noisy one, analogous to back-translation (Sennrich et al., 2016). Similarly, many other studies rely on the synthetic generation of noisy data for tasks such as grammatical error correction (Foster and Andersen, 2009), creating noise-resistant word embedding (Doval et al., 2019; Malykh et al., 2018) and machine translation (Bogoychev and Sennrich, 2019).

In comparison to the previous work, our work focuses on text anomalies caused by the usage of a different script in bilingual communities. In addition, we propose modeling the problem as a machine translation and generating synthetic data by script mapping. To the best of our knowledge, our approach to this problem has not been previously explored for the selected languages.

## 3 Methodology

This section presents our methodology to collect data, create script mapping to generate synthetic data and implement a transformer-based model. Source and dominant in this context respectively refer to the language of the original text and that of the dominant one used unconventionally.

### 3.1 Data Collection

As the first step, we collect data written in the conventional script of the selected languages. To that end, we create corpora based on Wikipedia dumps.[2] Since Wikipedia is not available for Gorani, we use Ahmadi (2020b)'s corpus for Gorani. Unlike the Latin script of Kurmanji for which there are corpora and Wikipedia, such as Pewan (Esmaili et al., 2013), there is no corpus for Kurmanji written in its Perso-Arabic script. Instead of relying on unreliable transliteration tools to convert the existing Kurmanji data, we crawl data from mainly news websites in the Iraqi Kurdistan for Kurmanji using the Perso-Arabic script.[3] It is worth mentioning that for Sorani Kurdish we use a large existing corpus (Ahmadi and Masoud, 2020) instead of the (smaller) Wikipedia dump.

We clean raw text by removing hyperlinks, email addresses, dates, non-relevant symbols and zero-width non-joiner (ZWNJ), if not systematically used in the script. Different types of numerals, namely Eastern Arabic <٠١٢٣٤٥٦٧٨٩>, Farsi <٠١٢٣٤٥۶٧٨٩> and Hindu-Arabic <0123456789>, are unified with the later ones for consistency. We also deal with script switching in some Wikipedia articles, particularly in Sindhi and Kashmiri, using regular expressions to only keep the Perso-Arabic data.

Following this, we extract vocabularies from the corpora based on a frequency list; depending on the size and quality of the data, we select words appearing with a minimum frequency in the range of 3 to 10. In addition to the vocabulary extracted from corpora, we also collect word lists and bilingual dictionaries in the source and target languages from other sources online. We consulted Wiktionary[4] for Azeri Turkish, Kashmiri, Mazanderani, Sindhi and Sorani without finding any such resources for the other languages. Additionally, the lexicographical data provided by Ahmadi et al. (2019) were used for Sorani.

### 3.2 Script Mapping

To simulate the process that leads to noisy data, we create script mappings that map characters in the conventional script of the source language to that of the dominant language. To do so, we define mapping rules between the scripts based on the orthographies of the languages, as in the compound characters <ئ> in Kurdish (composed of <ئ> (U+0626) and <ی> (U+06CE)) that appear so only at the beginning of a word and this can be mapped to either <ا> (U+0627) or the same character but with the diacritic *Kasrah* as <ا>. In addition, we take the closest candidates in the other script into account according to Unicode normalization as in <ک> (U+06A9) and <ك> (U+0643), and visual normalization, i.e. resemblance of the graphemes as in <ڎ> (U+068E) and <ذ> (U+0630). Table 2 shows a few mapping rules.

| Language | Unconventional script | Source | Target |
|---|---|---|---|
| Azeri Turkish | Persian | چ | چ |
| Sorani | Arabic | ز | ذ / ض / ظ / ز |
| Kashmiri | Urdu | آ | أ / ا |
| Sindhi | Urdu | ي | ے / ي / ی |

Table 2: An example of script mapping rules. In unconventional writing, we assume that a character in the source language can be mapped to one or more characters in the target script. '/' specifies different mapping possibilities.

Using the rule-based script mappings, we also determine words in the word lists and bilingual dictionaries that are potential translations and written similarly in the two scripts. We also consider removing diacritics, also known as *Harakat*, as they are not always included in writing. The following words are collected this way: 'برنج' ('rice') in Kurmanji and Persian, 'سویدی' ('Swedish') written with <ی> (U+06CC) in Sorani and 'سویدي' written with <ي> (U+064A) in Arabic, 'أمریکی' ('American') in Kashmiri and 'امریکی' in Urdu and, 'بۆرج' ('tower') in Azeri Turkish and 'برج' in Persian by removing <ۆ> (U+06C6). As such, we refer to the set of words collected as spelling pairs.

### 3.3 Character-alignment Matrix

Although script mapping based on rules and modifications is effective, especially to retrieve common words or words borrowed by the two languages, it may lead to potentially false friends or incorrect spelling pairs as well. Hence, to capture information based on the spelling pairs, we rely on the character alignment of words. To this end, we employ Needleman and Wunsch (1970)'s algorithm for sequence alignment that maximizes the number of matches between sequences, i.e. words,

---

[2] Dumps of December 1, 2022.
[3] This corpus will be released along with the other data.
[4] https://www.wiktionary.org

$a$ and $b$ with respect to the length of the two sequences. Therefore, we define the alignment matrix $D$ for each spelling pair by setting elements $i$ in $a$ and $j$ in $b$ according to the following:

$$D_{i,j} = max \begin{cases} D_{i-1,j-1} + d_{a_i b_j} \\ D_{i-1,j} + w \\ D_{i,j-1} + w \end{cases}$$

where $D_{i,j}$ is the score of character $i$ in the sequence $a$ and character $j$ in the sequence $b$, $d_{a_i b_j}$ denotes the similarity score of the characters at $i$ and $j$ (1 if similar and -1 otherwise) and, $w$ refers to the gap penalty which is set to -1. A gap penalty is a penalizing score for non-matching characters and is shown by '-' in our implementation. The matrix is initialized with $D_{0,0} = 0$. This algorithm is beneficial for our task as it considers the two sequences globally and allows back-tracing, hence useful to provide sequence matches. The following example shows the alignment of 'فیه‌تنام' ('Vietnam') in Sorani with the same word 'ویتنام', in Persian using this algorithm:

| ف | ی | ه | ت | ن | ا | م | فیه‌تنام |
|---|---|---|---|---|---|---|---|
| و | ی | - | ت | ن | ا | م | ویتنام |

Finally, we merge all the alignment matrices, i.e. $D$s, and create a character-alignment matrix for each pair of source and dominant languages. This matrix is normalized to have a unit norm. Furthermore, the mappings based on the rules of script mapping described in the previous subsection are appended to the matrix with a probability of 1. Alignments with a score < 0.1 are removed from the matrix due to the low probability of replacement. Depending on the score, a character can be aligned to more than one character in the dominant script.

### 3.4 Synthetic Data Generation

Given that there is a limited amount of data available for the selected languages, and that identifying languages in noisy setups is a challenging task, we have to rely on synthetic data generation. To that end, we first extract sentences of minimal and maximal lengths of 5 and 20 tokens (space-delimited) from the corpora described in §3.1. Then, we replace characters in the extracted sentences (clean data) with those in the character-alignment matrix. The replacement is done randomly with uniform sampling from the character alignment matrices to increase diversity. Depend-ing on the number of replacements, we also consider specific percentages of noise generation at 20%, 40%, 60%, 80% and 100%. We create a last dataset by merging all datasets with all levels of noise; this creates more noisy instances given the randomness of noise generation. The parallel data are then tokenized using regular expressions.

The number of parallel sentences and words per noise scale is provided in Appendix C.

### 3.5 Implementation and Evaluation

We use JoeyNMT (Kreutzer et al., 2019) to train transformer encoder-decoder models at the character level based on the degree of noise and language pairs. Using this model, our objective is to encode noisy data, i.e. synthetically noisy sentences, and decode normalized ones, i.e. the clean sentences in the collected datasets. We report the performance of the models by BLEU score (Papineni et al., 2002) and character $n$-gram F-score (chrF Popović, 2015), both calculated based on Sacre-BLEU (Post, 2018), along with sequence-level accuracy (Seq. Acc.), i.e. number of correct tokens in the hypothesis appearing in the same position as the reference divided by all tokens. Hyperparameter details are outlined in Appendix B.

We evaluate the performance of the trained models based on the noise scale. As a naive baseline system, we calculate the evaluation metrics for the parallel data without applying any normalization technique. Finally, we evaluate the effectiveness of our models in two downstream tasks: language identification and machine translation.

## 4 Results and Analysis

### 4.1 Script Normalization

As the first set of experiments, we evaluate the performance of the script normalization models on synthetic noisy data at all levels of noise. In Appendix, Table D.3 provides the results of all the models and compares them with the baseline, and examples of normalization by our models are provided in Table D.1. Furthermore, the performance of a few selected models is presented in Figure 2.

Although our models perform competitively in comparison to the baseline, performance is not identical across the datasets. By increasing the noise level from 20% to 100% making the source data harder to be normalized, a gradual decrease in performance is expected, and we indeed observe this for both the baseline and our model, but for most datasets (7 out of 12) the degradation for the naive baseline is more rapid and pronounced. Our model does seem to handle various levels of noise: in Sindhi, for instance, we get 75.1 BLEU score vs. 19.7 of the baseline (see bottom right $\text{SND}_{\text{URD}}^{100} \rightarrow \text{SND}$ in Figure 2).

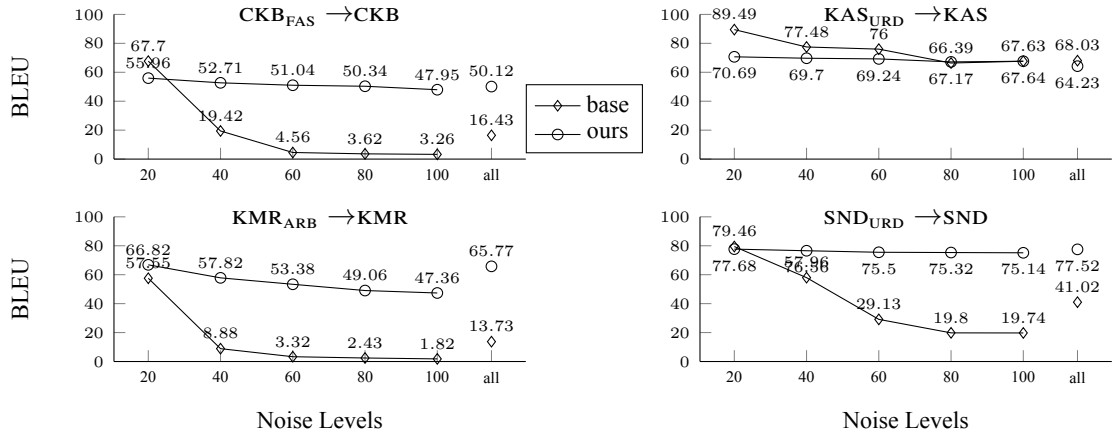

Figure 2: The performance of our models vs. the baseline in script normalization. See Figure D.1 for all languages.

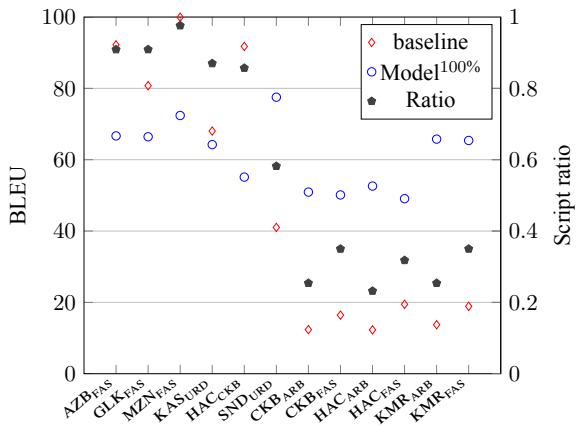

Figure 3: BLEU scores of the baseline and our model evaluated on 100% noisy data. The naive baseline outperforms our model for the settings where the differences between the noisy and "clean" data are minimal, i.e. when the script ratio (right y-axis) is high.

For five datasets, namely, $\text{AZB}_{\text{FAS}} \rightarrow \text{AZB}$, $\text{GLK}_{\text{FAS}} \rightarrow \text{GLK}$, $\text{MZN}_{\text{FAS}} \rightarrow \text{MZN}$, $\text{KAS}_{\text{URD}} \rightarrow \text{KAS}$ and $\text{HAC}_{\text{CKB}} \rightarrow \text{HAC}$, the naive baseline outperforms our models. We believe that this is explained by the level of similarity of the source and dominant scripts, which in turn determines the difficulty of script normalization. To quantify our assumption, we define $R_{A:B}$ as the script ratio of scripts A and B, both as two sets of characters, as follows:

$$R_{A:B} = \frac{A \cap B}{A \cup B} \times \frac{A \boxminus B}{A \cap B}$$

where $A \boxminus B$ refers to the intersection of characters in scripts $A$ and $B$ which are mapped in the rule-based script mapping (see §3.2) without any other alternative in the other script while $A \cap B$ is the intersection of $A$ and $B$ regardless of the mapping. Intuitively speaking, the script ratio of two identical scripts should be closer to 1 while more different scripts with various mappings between characters should get a lower value. Table A.1 provides the script ratios.

Figure 3 shows the BLEU score (left y-axis) of the baseline and our model to normalize the datasets containing 100% noise, e.g. $\text{GLK}_{\text{FAS}}^{100} \rightarrow \text{GLK}$ along with the script ratio for each language (right y-axis). This indicates that the normalization model (model$_{100}$) performs better where the script ratio is relatively low (<0.6, i.e. the two scripts are not that similar). On the other hand, the baseline performs better for higher script ratios, because in general the two scripts are very similar and hence there is less "noise". We leave for future work an exploration as to why our transformer models fail to even simply learn to copy their input to perform at least on par with the baseline.

**Real Data Evaluation** Given the scarcity of real data, we resorted to generating synthetic data for training models and consequently evaluating them. However, working with real data is also crucial to evaluate the effectiveness of our approach. As such, we collected 100 sentences from social media in Sorani Kurdish written in unconventional scripts of Persian and Arabic. These sentences are then manually normalized by native and expert speakers based on the standard orthography of Kurdish.

| Sorani Eval Set | Original (Unconventional) | | Normalized | |
|---|---|---|---|---|
| | BLEU | chrF | BLEU | chrF |
| CKB$_{\text{FAS}}$→CKB | 1.2 | 38.4 | 20.1 | 69.6 |
| CKB$_{\text{ARB}}$→CKB | 0.4 | 19.4 | 12.7 | 65.2 |

Table 3: Experiments on normalization of real-world data. The source sentences in Sorani Kurdish are written in the unconventional scripts of Persian (CKB$_{\text{FAS}}$) and Arabic (CKB$_{\text{ARB}}$). Even in this challenging setting (note how different the unconventional sentences are, as evidenced by low scores in the left column), model$_{100}$ manages to decently normalize them.

The results of the small-scale evaluation on the real data are provided in Table 3. In these datasets, calculating BLEU scores of the source sentences (noisy) with respect to the reference ones (clean) for CKB$_{\text{FAS}}$→CKB and CKB$_{\text{ARB}}$→CKB gets to 1.2 and 0.4 points, respectively. Once the source sentences are normalized using model$_{100}$, the corresponding BLEU scores increase to 20.1 for CKB$_{\text{FAS}}$→CKB and 12.7 for CKB$_{\text{ARB}}$→CKB. We selected this model as it has been trained on the most diverse training set. We believe that such a boost in BLEU scores indicates the robustness of our models to effectively normalize unconventional writing.

## 4.2 Language Identification

Language identification is the task of detecting the language of a text, usually a sentence. It is modeled as a probabilistic classification problem. As the first downstream task, we carry out a few experiments on language identification in three setups:

1. CLEAN: identifying languages without injecting any noise in the datasets, i.e. the target sentences in the parallel data. This is equivalent to 0% of noise in the data.
2. NOISY: identifying languages with noisy data at various levels, starting from 20% of noise and gradually increasing 20% up to 100%. We combine all data with all levels of noise in a separate dataset referred to as ALL.
3. MERGED: merging CLEAN with ALL dataset, i.e. with all noisy data.

We use the Tatoeba sentence dataset[5] for data in Persian, Urdu and Arabic, with additional data for Urdu from the TED corpus on Opus (Tiedemann, 2012). To tackle data imbalance, we downsampled all the datasets to only include 6000 sentences for each language[6]. In the MERGED setup where there are 12,000 sentences per language (half noisy, half clean), additional sentences (clean) in Persian, Urdu and Arabic are added to avoid data imbalance. Finally, we then split datasets into train and test sets with an 80-20% split. To train supervised language identification models, we use fastText (Bojanowski et al., 2017) with subword features with minimum and maximum character $n$-gram sizes of 2 and 4, word vectors of size 16 and hierarchical softmax as the loss function.

---
[5] https://tatoeba.org

| Noise % | Model | P@1 | R@1 | F@1 | P@2 | R@2 | F@2 |
|---|---|---|---|---|---|---|---|
| 0 | ours | 0.52 | 0.54 | 0.51 | 0.32 | 0.64 | 0.32 |
| | **fastText** | 0.69 | 0.69 | 0.69 | 0.39 | 0.78 | 0.39 |
| 20 | **ours** | 0.93 | 0.93 | 0.93 | 0.48 | 0.97 | 0.48 |
| | fastText | 0.71 | 0.71 | 0.71 | 0.44 | 0.89 | 0.44 |
| 40 | **ours** | 0.91 | 0.9 | 0.91 | 0.48 | 0.96 | 0.48 |
| | fastText | 0.56 | 0.56 | 0.56 | 0.38 | 0.76 | 0.38 |
| 60 | **ours** | 0.9 | 0.9 | 0.9 | 0.48 | 0.96 | 0.48 |
| | fastText | 0.41 | 0.41 | 0.41 | 0.31 | 0.63 | 0.31 |
| 80 | **ours** | 0.9 | 0.9 | 0.9 | 0.48 | 0.96 | 0.48 |
| | fastText | 0.37 | 0.37 | 0.37 | 0.28 | 0.57 | 0.28 |
| 100 | **ours** | 0.89 | 0.89 | 0.89 | 0.48 | 0.96 | 0.48 |
| | fastText | 0.37 | 0.37 | 0.37 | 0.28 | 0.57 | 0.28 |
| All | **ours** | 0.91 | 0.91 | 0.91 | 0.48 | 0.96 | 0.48 |
| | fastText | 0.48 | 0.48 | 0.48 | 0.34 | 0.68 | 0.34 |
| Merged | **ours** | 0.72 | 0.72 | 0.72 | 0.4 | 0.8 | 0.4 |
| | fastText | 0.69 | 0.69 | 0.69 | 0.4 | 0.79 | 0.4 |

Table 4: The performance of language identification using the pretrained fastText model as the baseline in comparison to our model trained on our datasets with various noise levels. Our model handles different levels of noise (rows 20 to All) and outperforms the baseline that is only trained on "clean" data. Models with the highest $F_1$ measure in first suggestions (F@1) are **bold**.

Table 4 presents the results of the performance of our models in comparison to fastText's language identification model trained on data from Wikipedia, Tatoeba and SETimes on 176 languages,[7] including, Persian, Arabic, Urdu, Sindhi, Sorani and Mazanderani. Although Azeri Turkish is supported, it is not clear which script it is trained on in fasText. The results are reported based on precision, recall and $F_1$ score of the first and second detection of the models, respectively denoted by '@1' and '@2'. Since Gorani and Gilaki are not among the Fairseq-supported languages, we focus our analysis on the top-two predictions (@2) to en-sure fairness against the baseline.

---
[6] Kashmiri had only 4700 instances.
[7] As of January 2023.

Our models outperform fastText in noisy conditions, regardless of the level of noise. The baseline performance degrades faster as the level of noise increases. Our models perform less effectively on clean data but recall that they are trained on a substantially smaller amount of data.[8]

| Gold Labels \ Model Predictions | Azeri | Gilaki | Mazandarani | Arabic | Persian | Gorani | Sorani | Kurmanji | Kashmiri | Sindhi | Urdu |
|---|---|---|---|---|---|---|---|---|---|---|---|
| Azeri | 1383 | 312 | 301 | 0 | 0 | 98 | 114 | 112 | 6 | 2 | 0 |
| Gilaki | 98 | 1116 | 107 | 1 | 1 | 75 | 98 | 83 | 6 | 26 | 0 |
| Mazandarani | 126 | 149 | 1199 | 1 | 19 | 85 | 89 | 70 | 18 | 11 | 0 |
| Arabic | 1 | 2 | 1 | 2368 | 10 | 3 | 0 | 0 | 0 | 4 | 5 |
| Persian | 0 | 5 | 13 | 10 | 2366 | 0 | 0 | 0 | 2 | 0 | 3 |
| Gorani | 215 | 222 | 199 | 9 | 4 | 1327 | 330 | 321 | 0 | 12 | 0 |
| Sorani | 320 | 327 | 325 | 3 | 0 | 488 | 1336 | 477 | 0 | 8 | 0 |
| Kurmanji | 170 | 170 | 156 | 0 | 0 | 239 | 301 | 1223 | 2 | 10 | 2 |
| Kashmiri | 4 | 4 | 5 | 2 | 0 | 2 | 5 | 5 | 2086 | 46 | 1 |
| Sindhi | 83 | 93 | 94 | 5 | 0 | 87 | 125 | 108 | 18 | 2271 | 5 |
| Urdu | 0 | 0 | 0 | 1 | 0 | 0 | 2 | 1 | 2 | 10 | 2383 |

Figure 4: Language identification using the MERGED model. The number of classifications is annotated.

Figure 4 illustrates the results of classification in our MERGED model where the detected languages (x-axis) are compared with the references (y-axis). Although the model performs well in detecting Arabic, Persian, Sindhi, Urdu and Kashmiri, the other languages are often confused. The misclassified languages can be categorized into two groups of [Azeri, Gilaki, Mazandarani] and [Sorani, Kurmanji, Gorani]. This can be explained by the similarity of scripts. Surprisingly, Sindhi and Kashmiri represent a less salient overlap in classification.

## 4.3 Neural Machine Translation

Furthermore, we evaluate the performance of the script normalization models in neural machine translation (NMT). To do so, we use the `devtest` data of the FLORES 200 dataset (Costa-jussà et al., 2022) that contains 1012 parallel sentences for 204 languages including Kashmiri, Sorani and Sindhi in their Perso-Arabic scripts along with English. With English as the target language and the other available languages as sources, we carry out the evaluation in the following three setups:

1. CLEAN: translating the `devtest` of the source languages as it is in the FLORES dataset (with no noise), e.g. CKB→ENG for translating Sorani (CKB) to English (FAS). This is a skyline setting (best possible).

2. NOISY: translating the `devtest` by synthetically applying a certain amount of noise. For this setting, we specify the maximum amount of "interference" with the dominant language's script (100% following the previous notation). The noise generation is random similar to language identification (§3.4). This is the baseline setting (worst possible, no normalization).

3. NORMALIZED: we normalize the data from the NOISY setting with our text normalization models (§4.1), and then translate the normalized text to English.

We use Facebook's No Language Left Behind (NLLB) model on HuggingFace[9] which is trained on parallel multilingual data from a variety of sources to translate into English and evaluate using SacreBLEU (Post, 2018). Table D.2 presents results of the translation quality of the NLLB model in our intended setups, with a sample in Table 5. Except for KAS_URD→ENG, the translation quality of unnormalized data (S) deteriorates as noise levels increase while it remains steady when using normalization models as a preprocessing step (H). In Sorani, for example, even with a 100% of noise (CKB$_{ARB}^{100}$), our normalization approach (H) recovers almost 20 BLEU points out of the 26 points that were lost due to the non-conventional setting. The importance of a trustworthy text normalization system is clear in the case of Kashmiri, where our normalization model fails to reduce noise, resulting in subpar performance.

From a qualitative point of view, the noisy input affects the quality of translations depending on the type of noise. Although the NLLB model shows resilience to certain types of errors, particularly where a character is substituted by an incorrect one without an impact on word boundaries, i.e. no spaces being added or removed, noisy characters that can possibly affect tokenization lead to incorrect translations. For instance in Table D.4, the translation of noisy sentences in Sindhi is less different from that of the reference, while the Sorani noisy sentence is translated terribly differently and incorrectly. Needless to say, the meaning of some words varies with incorrect characters.

---

[8]We conduct a similar study on a wider range of languages in (Ahmadi et al., 2023).

[9]The `nllb-200-distilled-600M` variant.

| Language | Noise % | Test set | BLEU | chrF |
|---|---|---|---|---|
| $\text{SND}^{100}_{\text{URD}}\!\rightarrow\!\text{ENG}$ | 0 | R | 30.81 | 55.96 |
| | 100 | S | 7.09 | 27.59 |
| | | H | 23.97 | 49.47 |
| $\text{CKB}^{100}_{\text{ARB}}\!\rightarrow\!\text{ENG}$ | 0 | R | 30.81 | 55.96 |
| | 100 | S | 4.72 | 24.33 |
| | | H | 24.53 | 50.55 |
| $\text{CKB}^{100}_{\text{FAS}}\!\rightarrow\!\text{ENG}$ | 0 | R | 29.69 | 45.79 |
| | 100 | S | 17.29 | 25.39 |
| | | H | 26.27 | 46.7 |
| $\text{KAS}^{100}_{\text{URD}}\!\rightarrow\!\text{ENG}$ | 0 | R | 28.40 | 55.71 |
| | 100 | S | 27.93 | 55.13 |
| | | H | 11.39 | 35.97 |

Table 5: Our normalization model –results with (H) data– largely (except for Kashmiri) mitigates the quality issues suffered from unconventional noisy inputs (S), almost to the levels of clean reference data (R).

## 5   Conclusion and Discussion

This paper discusses the challenge of script normalization of unconventional writing of languages that are spoken in bilingual communities. Under the influence of the dominant language of such communities, speakers tend to use scripts or orthographies of the dominant language rather than the one that is conventionally used in their native language. This leads to noisy data that hinder NLP applications and reduce data availability.

Framing the problem as a machine translation one where noisy data is 'translated' into clean one, we address the task of script normalization for the Perso-Arabic scripts of several languages, namely Azeri Turkish, Mazanderani, Gilaki, Sorani and Kurmanji Kurdish, Gorani, Kashmiri and Sindhi, with dominant languages being Arabic, Persian and Urdu. Given that these languages are less-resourced, we rely on synthetic data to create parallel datasets by injecting noise based on script mapping. We then train transformer-based encoder-decoders and show that the models can tackle the problem effectively, even for real data, but differently based on the level of noise.

We demonstrate that script normalization can be effectively addressed, particularly where there are many dissimilarities between the source and dominant languages. It can also alleviate the impact of noisy data on downstream tasks, namely language identification and machine translation. In addition to script normalization that can help retrieve texts for the selected under-resourced languages, the trained models along with the collected data, with or without noise, can pave the way for further developments for those languages.

**Limitations**   One of the limitations of the current study is the lack of annotated data for all languages. This is also the case of machine translation for which data could only be found for Kashmiri, Sorani and Sindhi, while other languages do not have much parallel data yet. On the other hand, the notion of noisy data is limited to the replacement of the missing characters in a script when compared to another one, i.e. that of the dominant language. As an ablation study, injecting other types of noise, beyond those discussed in this paper, may improve the performance of the models to tackle not only script normalization but several related tasks such as spelling error correction and may also increase the robustness of the models for morphologically rich languages or languages with versatile word boundaries using ZWNJ. Although we did our best to filter out code-switched data in the corpora, our datasets may contain data in other languages (in Perso-Arabic scripts).

In future work, we would like to apply our approach to other scripts and languages in bilingual communities. We also suggest evaluating the impact of script normalization on more downstream tasks, especially transliteration and tokenization.

## Ethics Statement

All corpora and datasets used in this study are publicly available, ensuring compliance with data privacy regulations. Although we did our best to remove any personally identifiable information and preserve the privacy and anonymity of individuals, it is possible that some of the selected corpora contain sensible or offensive information. Filtering such content is challenging given that all the targeted languages are low-resourced and lack proper NLP tools for this purpose. We believe that it is unlikely that the normalization models cause benefits or harm to individuals. Regarding the annotation of the data (§4.1), annotators were fairly compensated for their time and effort. By upholding these ethical principles, we aimed to conduct the study in a responsible and conscientious manner.

## Acknowledgments

This work is supported by the National Science Foundation under DLI-DEL award BCS-2109578. The authors are also grateful to the anonymous reviewers, as well as the Office of Research Computing at GMU, where all computational experiments were conducted.

## A  Selected Languages

In this work, we focus on script normalization for a few languages that are spoken in bilingual communities and use a Perso-Arabic script.

Even though the Perso-Arabic script is not limited to our selected languages, we did not include languages that are spoken in countries where the administratively dominant language uses another script, such as Uyghur and Malay, or those that have historically used a Perso-Arabic script that is now obsolete, like Dogri, Turkish and Tajik Persian. This said, there are other languages that would fit into our study but due to lack of data could not be included, such as Luri (LDD), Balochi, Shina, or Burushaski.

**Azeri Turkish** also known as Azerbaijani or Azari, is a Turkic language mainly spoken in Azerbaijan, Iranian Azerbaijan and broadly the Caucasus by 20 million speakers (Rezaei et al., 2017). The two varieties of Azeri Turkish, Northern Azeri Turkish and Southern Azeri Turkish respectively use the Latin and the Perso-Arabic scripts. In this study, we focus on the latter variety. The Perso-Arabic script of Azeri Turkish is similar to that of Persian, with additional graphemes such as <ئی> (U+063D) and <ﺝ> (U+06C7).

**Mazanderani** also known as Mazandarani or Tabari (Borjian and Borjian, 2023), is an Indo-Aryan language spoken in regions adjoining the Caspian Sea in Iran, chiefly in Mazandaran Province, by 2.5-3 million speakers (Mirhosseini, 2015, p. 157). The Perso-Arabic script adopted for Mazandarani is almost identical to the one used for Persian, with the additional diacritic <ˇ> (U+02C7).

**Gilaki** is an Indo-Aryan language, similar to Mazanderani spoken in regions adjoining the Caspian Sea in Iran by over 2 million speakers (Rastorgueva et al., 2012; Khoshrouz, 2021). Although Mazanderani and Gilaki have been historically written using the Persian script (Borjian, 2008) without a recognized orthography, there have been recently many efforts, particularly on Wikipedia[10], to adopt a Latin-based script. This said, the Latin script did not survive and the Perso-Arabic scripts are used for both Gilaki and Mazanderani (Sedighi, 2023, p. 20). The Perso-Arabic script adopted for Gilaki has graphemes in addition to those in Persian, such as <ﻮ> (U+06CB) and <ﻭ> (U+06CA).

**Kurdish** is an Indo-Aryan language spoken by over 25 million speakers in Iraq, Iran, Turkey and Syria, in varieties classified as Northern Kurdish or Kurmanji, Central Kurdish or Sorani, and Southern Kurdish (Ahmadi and Masoud, 2020). Kurdish has historically employed various scripts for writing, such as Cyrillic, Armenian, Latin and Perso-Arabic. Among these, the two latter scripts are still widely used. While the Latin script is more popular for writing Kurmanji spoken in Turkey, speakers of other varieties prefer the Perso-Arabic script of Kurdish, mainly due to the widespread usage of those scripts by the regional administrations. This is particularly the case of Sorani and Kurmanji spoken in Iraq and to some extent, in Syria. The Kurmanji variant spoken in Kurdistan of Iraq is called Behdini (also spelled Badini). Kurdish is a phonemic alphabet which is known by its distinct characters such as <ۆ> (U+06C6) and <ێ> (U+06CE), and absence of varying phonetically-related graphemes of Arabic such as <ض> (U+0636) and <ظ> (U+0638) even for loanwords (Ahmadi, 2020a).

It is worth noting that the Perso-Arabic script that is used for both Sorani and Kurmanji follow the same orthographies. However, this is less established for Kurmanji given the popularity of the Latin script for that dialect. Throughout the paper, we consider Kurmanji and Sorani differently due to the considerable lexical and morpho-syntactic differences. It is worth noting that Gorani speakers oftentimes use Sorani script without the additional characters of Gorani making Sorani a dominant language.

**Gorani** also known as Hawrami, is an Indo-Aryan language spoken by 300,000 speakers in the parts of the Iranian of Iraqi Kurdistan (Ahmadi, 2020b). Due to the mutual linguistic and cultural influences of Sorani and Gorani, Gorani speakers rely on the Perso-Arabic script and the orthography used for Kurdish. However, there are a few additional graphemes that are unique to Gorani and cannot be found in Kurdish, such as <ڎ> (U+068E) and <ﻮ> (U+06CB).

**Kashmiri** also known as Koshur, is an Indo-Aryan language spoken by over 7 million speakers in the disputed territories administered by three countries: India, Pakistan, and China (Lone et al.,

---

2022b). Although Kashmiri's Perso-Arabic script relies much on that of Urdu, the extensive use of diacritics and distinct characters such as <ۄ> (U+06C4) and <ؠ> (U+0620) make the adopted script discernible. Moreover, Kashmiri extensively uses diacritics to specify vowels, while diacritization is less frequent in everyday writing of the other languages, except for disambiguation.

**Sindhi** is an Indo-Aryan language primarily spoken in Pakistan and parts of India by over 20 million speakers (David et al., 2017). Although influenced by Urdu, Sindhi uses a more diverse set of letters (62 vs. 40 in Urdu) (Doctor et al., 2022). It can be distinguished by unique letters such as <ڏ> (U+068F) and <ڻ> (U+06BB).

| Language | Script ratio |
|---|---|
| MZN$_{FAS}$ | 0.976 |
| AZB$_{FAS}$ | 0.909 |
| GLK$_{FAS}$ | 0.909 |
| KAS$_{URD}$ | 0.87 |
| HAC$_{CKB}$ | 0.857 |
| SND$_{URD}$ | 0.582 |
| CKB$_{FAS}$ | 0.35 |
| KMR$_{FAS}$ | 0.35 |
| HAC$_{FAS}$ | 0.318 |
| CKB$_{ARB}$ | 0.254 |
| KMR$_{ARB}$ | 0.254 |
| HAC$_{ARB}$ | 0.232 |

Table A.1: Script ratios as defined in §4.1

Similar to Persian and Urdu, Azeri Turkish, Mazandarani and Gilaki use the zero-width nonjoiner (ZWNJ, U+200C) character (Lockwood, 2011) which creates lexical variations and adds to the complexity of downstream tasks like tokenization (Ghayoomi and Momtazi, 2009; Rehman et al., 2011; Doostmohammadi et al., 2020). Furthermore, the usage of some graphemes such as <ض> (U+0636) and <ظ> (U+0638) in the adopted scripts is mainly limited to loanwords from Arabic and Persian, unless conventionally used to represent a specific phoneme in the language. Moreover, the final glyph <ھ> (U+06BE) is used in all languages except Persian, Mazandarani, Azeri Turkish and Gilaki. While the Perso-Arabic script of Kurdish, Gorani and Kashmiri are alphabets, the other selected languages are impure Abjads. Similarly, all the selected languages use Naskh style for writing, unlike Urdu that uses Nastaliq.

A summary of various aspects of the selected languages is provided in Table B.1. We also provide the script ratio (as defined in §4.1) of the script of the selected languages and that of the dominant one in Table A.1.

## B Models Details

We train the normalization models in different configuration of hyper-parameters as follows:
- Training:
  - level: character
  - maximum number of characters: 100
  - character minimal frequency: 5
  - optimizer: adam
  - learning rate: 0.001
  - learning rate min: 0.0002
  - patience: 5
  - number of layers: 6
  - number of heads: 8
  - embedding dimension: 128
  - hidden size: 128
  - position-wise feed-forward layer size: 512
- Testing:
  - number of best prediction: 1
  - beam size: 4
  - beam alpha: 1.0
  - max output length: 100
  - batch size: 10

In addition to these common hyper-parameters, we set different hyper-parameters depending on the size of the datasets. The approximate overall number of pairs in all noisy datasets are provided in parentheses.
- (<10k pairs) KAS$_{URD}$, HAC$_{CKB}$
  epochs: 100 / dropout: 0.2 / batch size: 64
- (<50k pairs) KMR$_{ARB}$, HAC$_{ARB}$, KMR$_{FAS}$, HAC$_{FAS}$, GLK$_{FAS}$, MZN$_{FAS}$
  epochs: 80 / dropout: 0.2 / batch size: 64
- (<600k pairs) SND$_{URD}$, AZB$_{FAS}$
  epochs: 40 / dropout: 0.1 / batch size: 64
- (<6.6M pairs) CKB$_{ARB}$, CKB$_{FAS}$
  epochs: 6 / dropout: 0.1 / batch size: 10

For each configuration (72 in total), the training was carried out on a cluster using one GPU per configuration with 32GB memory.

## C Datasets

The number of words and sentence pairs in the synthetic datasets generated for the selected languages is provided in Table C.1.

| Language | 639-3 | WP | script type | diacritics | ZWNJ | Dominant |
|---|---|---|---|---|---|---|
| Azeri Turkish | `azb` | `azb` | Abjad | ✓ | ✓ | Persian |
| Kashmiri | `kas` | `ks` | Alphabet | ✓ | ✗ | Urdu |
| Gilaki | `glk` | `glk` | Abjad | ✓ | ✓ | Persian |
| Gorani | `hac` | – | Alphabet | ✗ | ✗ | Persian, Arabic, Sorani |
| Kurmanji | `kmr` | – | Alphabet | ✗ | ✗ | Persian, Arabic |
| Sorani | `ckb` | `ckb` | Alphabet | ✗ | ✗ | Persian, Arabic |
| Mazanderani | `mzn` | `mzn` | Abjad | ✓ | ✓ | Persian |
| Sindhi | `snd` | `sd` | Abjad | ✓ | ✗ | Urdu |
| Persian | `fas` | `fa` | Abjad | ✓ | ✓ | - |
| Arabic | `arb` | `ar` | Abjad | ✓ | ✗ | - |
| Urdu | `urd` | `ur` | Abjad | ✓ | ✓ | - |

Table B.1: Selected languages studied in this paper. Columns 2 and 3 show the codes of the languages in ISO 639-3 and on their specific Wikipedia (WP). Diacritics refers to the usage of diacritics (Harakat) as individual characters, dominant is the administratively dominant language in the bilingual community where the language is spoken.

| SRC-TRG | | Noise % | | | | | |
|---|---|---|---|---|---|---|---|
| | | 20 | 40 | 60 | 80 | 100 | All |
| AZB_FAS | Pairs | 517860 | 517860 | 517860 | 517860 | 517860 | 584229 |
| | Words | 4266950 | 4266950 | 4266950 | 4266950 | 4266950 | 4987065 |
| CKB_ARB | Pairs | 1220386 | 1326715 | 1411998 | 1441641 | 1451201 | 6663362 |
| | Words | 13522986 | 14554313 | 15183038 | 15381207 | 15457911 | 72697912 |
| CKB_FAS | Pairs | 1186567 | 1325960 | 1408885 | 1435023 | 1442000 | 6491240 |
| | Words | 12838133 | 14381425 | 15125839 | 15305707 | 15363441 | 70402972 |
| GLK_FAS | Pairs | 16779 | 16779 | 16779 | 16779 | 16779 | 22215 |
| | Words | 176602 | 176602 | 176602 | 176602 | 176602 | 240833 |
| HAC_ARB | Pairs | 4718 | 4767 | 4802 | 4805 | 4802 | 23398 |
| | Words | 49804 | 50244 | 50457 | 50476 | 50464 | 248025 |
| HAC_CKB | Pairs | 4668 | 4669 | 4672 | 4686 | 4687 | 6474 |
| | Words | 49191 | 49195 | 49218 | 49417 | 49424 | 71246 |
| HAC_FAS | Pairs | 4712 | 4773 | 4798 | 4803 | 4802 | 23104 |
| | Words | 49646 | 50232 | 50429 | 50469 | 50464 | 244911 |
| KAS_URD | Pairs | 4729 | 4729 | 4734 | 4761 | 4759 | 9463 |
| | Words | 43907 | 43907 | 43925 | 44060 | 44064 | 96159 |
| KMR_ARB | Pairs | 10659 | 10963 | 11334 | 11417 | 11412 | 54430 |
| | Words | 96441 | 98403 | 100463 | 100877 | 100874 | 490034 |
| KMR_FAS | Pairs | 10631 | 10997 | 11336 | 11420 | 11417 | 53272 |
| | Words | 95971 | 98474 | 100454 | 100906 | 100884 | 482298 |
| MZN_FAS | Pairs | 36663 | 36663 | 36663 | 36663 | 36663 | 36665 |
| | Words | 365428 | 365428 | 365428 | 365428 | 365428 | 365446 |
| SND_URD | Pairs | 122446 | 122537 | 122865 | 122908 | 122946 | 496239 |
| | Words | 1328696 | 1329684 | 1333815 | 1334417 | 1334770 | 5581407 |

Table C.1: Number of words and aligned sentences (pairs) in the synthetic datasets

# D Experiments Results and Examples

This section presents examples of script normalization and machine translation. Given the source sentence, the model trained on all noisy datasets, i.e. All, generates a hypothesis which is then translated. The selected sentences are taken from the 100% noisy dataset. The source, hypothesis and reference sentences are specified as S, H and R, respectively.

| Language | S/H/R | Sentence |
|---|---|---|
| CKB$_{ARB}$→CKB | S | دزی حذبە دەمسڵاطدار و براوەکانی کوردسطان کار دەکەن |
| | H | دزی حزبە دەمسڵاتدار و براوەکانی کوردستان کار دەکەن |
| | R | دزی حزبە دەمسڵاتدار و براوەکانی کوردستان کار دەکەن |
| GLK$_{FAS}$→GLK | S | کلآحمد شنه عزیز خۊنه و نگارۀ خۏش همرأ بئنه |
| | H | کلآحمد شنه عزیزّ خۊنه و نگارۀ خۏشّ همرأ بئنه |
| | R | کلآحمد شنه عزیزّ خۊنه و نگارۀ خۏشّ همرأ بئنه |
| HAC$_{FAS}$→HAC | S | یاگی ویشن ئی زوانانی ب نیم مرد نامیشان بنیمی |
| | H | یاگی ویشمنه ئی زوانانی به نیمه مەردە نامیشان بنیمی |
| | R | یاگی ویشمنه ئی زوانانیه به نیمه مەردە نامیشان بنیمی |
| KAS$_{URD}$→KAS | S | رامہ بۆنے سبز ماد چھیے رامہ بۆنے |
| | H | رامہ بۆنۣی سبز ماد چھیے رامہ بۆنۣی |
| | R | رامہ بۆنۣی سبز ماد چھیے رامہ بۆنۣی |
| KMR$_{FAS}$→KMR | S | دلن والا ژ کەمرب و کینه نه دەرد و کول نه ژی کەمسەر |
| | H | دلین ڤالا ژ کەمرین و کینه نه دەرد و کول نه ژی کەمسەر |
| | R | دلین ڤالا ژ کەمرین و کینه نه دەرد و کول نه ژی کەمسەر |
| SND$_{URD}$→SND | S | کاٹائٽ علم کیمیا یا کیمسٽری م ھڪ مرکّب جو نالو آٻی |
| | H | حقاٽائٽ علم کیمیا یا کیمسٽری م ھڪ مرڪّب جو ناٺو آھي |
| | R | حقاٽائٽ علم کیمیا یا کیمسٽری م ھڪ مرڪّب جو ناٺو آھي |

(a) Script normalization of sentences containing 100% noise

| Source | شاری اہ صطوہ کان، ب باشطرین گرانی رہ صدہ نا پاٹوران |
|---|---|
| Reference | شاری ٺەسٹیۆکان، بۆ باشترین گۆرانی رەسمان پاٹیوران |
| | City of Stars, received nominations for best original song |
| 20 | شاری اہ صطوہ کان، ب باشطرین گرانی رہ صدہ نا پاٹوران |
| 40 | شاری ٺەسٹۆکان، بۆ باشترین گۆرانی رەسمان پاٹیوران |
| 60 | شاری ٺەسٹۆکان، بۆ باشترین گۆرانی رەسمان پاٹیورانشا |
| 80 | شاری ٺەسٹیۆکان، بۆ باشترین گۆرانی رەسمان پاٹیوران |
| 100 | شاری ٺەسٹیۆۆ ٘ان، بۆ باشترین گۆرانی رە سە نا پاٹیۆر |
| All | شاری ٺەسٹیۆکان، بۆ باشترین گۆرانی رەسمان پاٹیوران |

(b) Script normalization of a Sorani sentence with varying noise

Table D.1: Examples of script normalization of sentences in different languages containing 100% noise (a) and in Sorani Kurdish with various levels of noise (b) using a model trained on all noisy datasets

| Language | Noise % | Test set | BLEU | chrF |
|---|---|---|---|---|
| KAS$_{URD}^{100}$→ENG | 0 | R | 28.40 | 55.71 |
| | 20 | S | 28.42 | 55.70 |
| | | H | 11.60 | 36.15 |
| | 40 | S | 27.86 | 55.22 |
| | | H | 11.38 | 36.03 |
| | 60 | S | 27.88 | 55.14 |
| | | H | 11.38 | 36.05 |
| | 80 | S | 27.94 | 55.23 |
| | | H | 11.31 | 35.96 |
| | 100 | S | 27.93 | 55.13 |
| | | H | 11.39 | 35.97 |
| CKB$_{ARB}^{100}$→ENG | 0 | R | 30.81 | 55.96 |
| | 20 | S | 21.54 | 46.63 |
| | | H | 24.52 | 50.43 |
| | 40 | S | 15.16 | 39.49 |
| | | H | 21.76 | 47.49 |
| | 60 | S | 8.88 | 31.05 |
| | | H | 19.87 | 45.73 |
| | 80 | S | 4.75 | 24.59 |
| | | H | 24.11 | 50.06 |
| | 100 | S | 4.72 | 24.33 |
| | | H | 24.53 | 50.55 |
| CKB$_{FAS}^{100}$→ENG | 0 | R | 29.69 | 45.79 |
| | 20 | S | 21.04 | 39.95 |
| | | H | 23.27 | 38.07 |
| | 40 | S | 17.61 | 29.14 |
| | | H | 17.29 | 30.41 |
| | 60 | S | 21.04 | 38.56 |
| | | H | 26.31 | 44.04 |
| | 80 | S | 9.98 | 25.43 |
| | | H | 19.45 | 35.45 |
| | 100 | S | 17.29 | 25.39 |
| | | H | 26.27 | 46.7 |
| SND$_{URD}^{100}$→ENG | 0 | R | 30.81 | 55.96 |
| | 20 | S | 23.15 | 48.29 |
| | | H | 23.57 | 49.32 |
| | 40 | S | 18.03 | 42.60 |
| | | H | 21.84 | 47.57 |
| | 60 | S | 12.42 | 35.53 |
| | | H | 19.43 | 45.23 |
| | 80 | S | 6.95 | 27.45 |
| | | H | 23.39 | 49.51 |
| | 100 | S | 7.09 | 27.59 |
| | | H | 23.97 | 49.47 |

Table D.2: Evaluation of NLLB model for translation of reference data (R), noisy data (S) and normalized data (H) using normalization models trained on all levels of noise (All)

| Language | Noise % | Baseline | | | Our models | | |
|---|---|---|---|---|---|---|---|
| | | BLEU | chrF | seq. acc. | BLEU | chrF | seq. acc. |
| AZB$_{FAS}$ → AZB | 20 | 95.99 | 98.98 | 87.47 | 67.37 | 0.77 | 54.15 |
| | 40 | 92.37 | 98.05 | 79.51 | 67 | 0.77 | 54.01 |
| | 60 | 91.73 | 97.86 | 77.51 | 66.94 | 0.77 | 53.87 |
| | 80 | 91.53 | 97.8 | 76.69 | 66.89 | 0.76 | 53.67 |
| | 100 | 91.52 | 97.79 | 76.61 | 66.86 | 0.76 | 53.62 |
| | All | 92.2 | 97.95 | 77.41 | 66.67 | 0.76 | 51.75 |
| GLK$_{FAS}$ → GLK | 20 | 90.36 | 96.59 | 77.77 | 67.33 | 0.77 | 45.17 |
| | 40 | 82.93 | 93.16 | 65.91 | 67.4 | 0.77 | 44.87 |
| | 60 | 81.38 | 92.37 | 63.29 | 66.94 | 0.77 | 43.92 |
| | 80 | 80.99 | 92.11 | 61.92 | 66.47 | 0.77 | 42.61 |
| | 100 | 80.95 | 92.11 | 61.8 | 66.76 | 0.77 | 42.55 |
| | All | 80.75 | 92.22 | 58.51 | 66.43 | 0.77 | 38.16 |
| HAC$_{ARB}$ → HAC | 20 | 55.47 | 83.92 | 13.35 | 47.1 | 0.65 | 28.39 |
| | 40 | 10.44 | 46.13 | 1.05 | 38.95 | 0.63 | 17.4 |
| | 60 | 1.35 | 28.31 | 0 | 31.23 | 0.59 | 8.52 |
| | 80 | 1.53 | 18.88 | 0 | 25.16 | 0.56 | 7.28 |
| | 100 | 0.81 | 17.27 | 0 | 26 | 0.56 | 9.36 |
| | All | 12.29 | 38.79 | 2.39 | 52.61 | 0.68 | 27.69 |
| HAC$_{FAS}$ → HAC | 20 | 68.74 | 88.98 | 35.17 | 54.88 | 0.69 | 31.14 |
| | 40 | 17.75 | 53.16 | 7.32 | 33.65 | 0.6 | 14.02 |
| | 60 | 3.32 | 35.12 | 0.42 | 26.43 | 0.57 | 8.96 |
| | 80 | 2.71 | 28.12 | 0.42 | 23.05 | 0.54 | 6.86 |
| | 100 | 2.24 | 26.41 | 0.42 | 21.52 | 0.53 | 5.82 |
| | All | 19.46 | 46.84 | 8.09 | 49.07 | 0.66 | 25.14 |
| HAC$_{CKB}$ → HAC | 20 | 99.97 | 99.98 | 99.79 | 62.65 | 0.72 | 41.54 |
| | 40 | 99.44 | 99.75 | 97 | 62.24 | 0.72 | 40.26 |
| | 60 | 98.81 | 99.51 | 95.51 | 60.67 | 0.71 | 37.18 |
| | 80 | 93.04 | 97.83 | 78.46 | 60.04 | 0.71 | 38.38 |
| | 100 | 92.55 | 97.66 | 75.48 | 59.14 | 0.71 | 37.53 |
| | All | 91.75 | 97.33 | 71.91 | 55.12 | 0.68 | 30.56 |
| KAS$_{URD}$ → KAS | 20 | 89.49 | 96.91 | 71.25 | 70.69 | 0.81 | 49.89 |
| | 40 | 77.48 | 93.28 | 53.07 | 69.7 | 0.81 | 49.05 |
| | 60 | 76 | 92.65 | 48.1 | 69.24 | 0.81 | 46.41 |
| | 80 | 66.39 | 88.28 | 40.04 | 67.17 | 0.8 | 44.23 |
| | 100 | 67.63 | 88.77 | 41.18 | 67.64 | 0.8 | 46.22 |
| | All | 68.03 | 89.28 | 33.05 | 64.23 | 0.77 | 35.16 |
| MZN$_{FAS}$ → MZN | 20 | 99.93 | 99.98 | 99.86 | 72.38 | 0.8 | 50.97 |
| | 40 | 99.93 | 99.98 | 99.86 | 72.38 | 0.8 | 50.97 |
| | 60 | 99.93 | 99.98 | 99.86 | 72.38 | 0.8 | 50.97 |
| | 80 | 99.93 | 99.98 | 99.86 | 72.39 | 0.8 | 50.97 |
| | 100 | 99.93 | 99.98 | 99.86 | 72.39 | 0.8 | 50.97 |
| | All | 99.93 | 99.98 | 99.86 | 72.39 | 0.8 | 51 |
| KMR$_{ARB}$ → KMR | 20 | 57.55 | 82.52 | 12.1 | 66.82 | 0.78 | 43.43 |
| | 40 | 8.88 | 45.55 | 0.46 | 57.82 | 0.75 | 34.18 |
| | 60 | 3.32 | 31.48 | 0 | 53.38 | 0.73 | 27.78 |
| | 80 | 2.43 | 21.68 | 0 | 49.06 | 0.71 | 24.69 |
| | 100 | 1.82 | 19.81 | 0 | 47.36 | 0.7 | 24.61 |
| | All | 13.73 | 39.91 | 2.35 | 65.77 | 0.77 | 43.28 |
| KMR$_{FAS}$ → KMR | 20 | 69.98 | 88.35 | 27.26 | 68.92 | 0.79 | 45.39 |
| | 40 | 16.02 | 51.11 | 6 | 54.81 | 0.73 | 34.73 |
| | 60 | 5.5 | 36.06 | 0.18 | 50.11 | 0.72 | 26.1 |
| | 80 | 3.95 | 30.62 | 0.09 | 49.19 | 0.72 | 23.82 |
| | 100 | 3.52 | 28.72 | 0.18 | 45.11 | 0.7 | 23.73 |
| | All | 18.9 | 46.12 | 5.48 | 65.39 | 0.77 | 42.7 |
| CKB$_{ARB}$ → CKB | 20 | 55.89 | 84.47 | 9.15 | 54.54 | 0.66 | 30.93 |
| | 40 | 12.92 | 51.51 | 0.35 | 53.85 | 0.66 | 29.44 |
| | 60 | 3.27 | 31.48 | 0.04 | 53.38 | 0.66 | 28.33 |
| | 80 | 2.17 | 19.3 | 0.04 | 51.92 | 0.66 | 26.46 |
| | 100 | 2.14 | 17.36 | 0.04 | 50.11 | 0.65 | 26.16 |
| | All | 12.37 | 39.31 | 1.72 | 50.92 | 0.65 | 25.17 |
| CKB$_{FAS}$ → CKB | 20 | 67.7 | 89.45 | 26.9 | 55.96 | 0.67 | 33.49 |
| | 40 | 19.42 | 55.9 | 6.3 | 52.71 | 0.66 | 28.97 |
| | 60 | 4.56 | 35.67 | 0.2 | 51.04 | 0.65 | 26.89 |
| | 80 | 3.62 | 29.04 | 0.12 | 50.34 | 0.65 | 25.14 |
| | 100 | 3.26 | 26.43 | 0.14 | 47.95 | 0.64 | 24.44 |
| | All | 16.43 | 44.73 | 5.34 | 50.12 | 0.65 | 25.34 |
| SND$_{URD}$ → SND | 20 | 79.46 | 89.83 | 41.54 | 77.68 | 0.84 | 51.4 |
| | 40 | 57.96 | 78.51 | 14.91 | 76.56 | 0.83 | 48.57 |
| | 60 | 29.13 | 57.93 | 4.72 | 75.5 | 0.83 | 46.37 |
| | 80 | 19.8 | 49.49 | 3.52 | 75.32 | 0.83 | 46.24 |
| | 100 | 19.74 | 49.38 | 3.22 | 75.14 | 0.82 | 46.06 |
| | All | 41.02 | 64.43 | 11.15 | 77.52 | 0.83 | 48.79 |

Table D.3: Comparison of the results of our script normalization models vs. the baseline. 'All' refers to a model trained and tested on the merged datasets (20% to 100%) as a separate case.

| Language | Test$_{\text{Noise\%}}$ | Sentence |
|---|---|---|
| | R$_{\text{SND}}$ | ساڳئي ھائيڪنگ روٽ وانگر اسڪائنگ روٽ کي سمجهو. |
| | R$_{\text{ENG}}$ | Think of the skiing route as of a similar hiking route. |
| SND$_{\text{FAS}}{\rightarrow}$ENG | S$_{100\%}$ | ساڳئي ھائڪنگ روٽ وانگر اسڪائنگ روٽ کي سمجهو. |
| | T$_{100\%}$ | Think of a skiing route as the same as a hiking route. |
| | $\hat{S}_{100\%}$ | ساڳئي ھائيڪنگ روٽ وانگر اسڪائنگ روٽ کي سمجهو. |
| | $\hat{T}_{100\%}$ | Think of a skiing route as the same as a hiking route. |
| | R$_{\text{CKB}}$ | بیر له ڕێڕەوی خلیسکێنەی سەر بەفر بکەوە بۆ ڕێڕەوێکی ھاوشێوەی سەر شاخ کەوتن. |
| | R$_{\text{ENG}}$ | Think of the skiing route as of a similar hiking route. |
| CKB$_{\text{ARB}}{\rightarrow}$ENG | S$_{100\%}$ | بیر لة ڕرة ي خلیسکنة ي سة ر بة فر بکة ت ـ ب ررة کي ھاشة ي سة ر شاخ کة طن. |
| | T$_{100\%}$ | The church is divided into two parts, each of which is a mountain. |
| | $\hat{S}_{100\%}$ | بیر له ڕێڕەوی خولیسکۆنەی سەر بەفر پێکەوە بۆ ڕێڕەکی ھاوشێوەی سەر شاخ کەوتن. |
| | $\hat{T}_{100\%}$ | They thought of the snow-capped roller coaster route together to a similar route on the mountain. |

Table D.4: Examples of translating sentences into English using the NLLB model given a reference (R), a noisy sentence (S) and a normalized sentence ($\hat{S}$). Translations are shown with T.

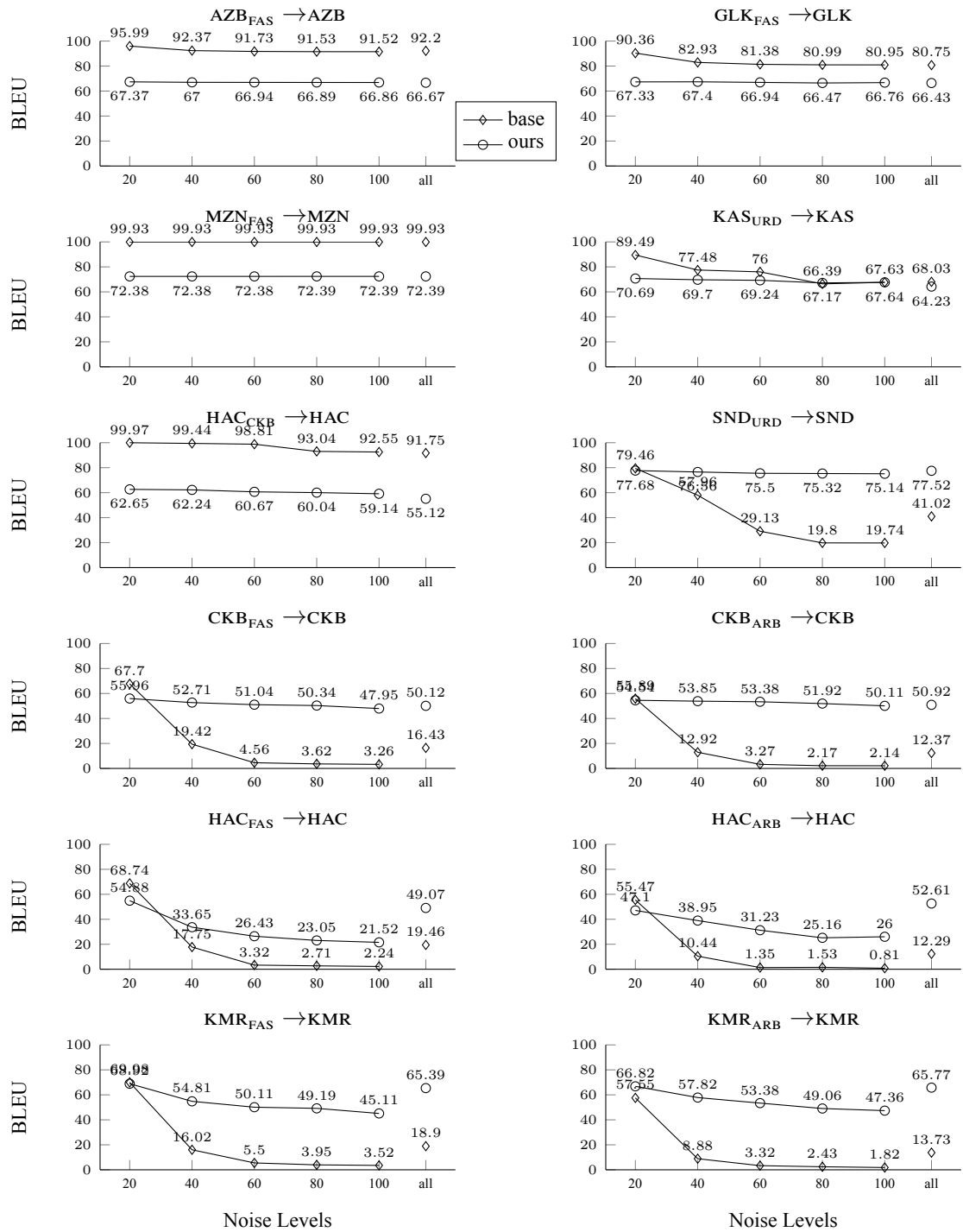

Figure D.1: Comparison of our normalization models to the naive "copy" baseline. In most languages, our normalization models dramatically improve over the baseline.